\documentclass[10pt,twocolumn,letterpaper]{article}

\usepackage{cvpr}
\usepackage{times}
\usepackage{epsfig}
\usepackage{graphicx}
\usepackage{amsmath}
\usepackage{amssymb}

\usepackage[breaklinks=true,bookmarks=false]{hyperref}
\usepackage[toc,page]{appendix}

\cvprfinalcopy 

\def\cvprPaperID{663} 

\ifcvprfinal\fi
\begin{document}

\title{Learning to Read Chest X-Rays:\\Recurrent Neural Cascade Model for Automated Image Annotation}

\author{Hoo-Chang Shin\textsuperscript{1}, Kirk Roberts\textsuperscript{2}, Le Lu\textsuperscript{1}, Dina Demner-Fushman\textsuperscript{2}, Jianhua Yao\textsuperscript{1}, Ronald M Summers\textsuperscript{1}\\
\\
\vspace{.32cm}
\begin{minipage}{\dimexpr0.42\linewidth-2\fboxsep-2\fboxrule\relax}
        \textsuperscript{1}Imaging Biomarkers and Computer-Aided Diagnosis Laboratory, Radiology and Imaging Sciences, Clinical Center
\end{minipage}\hspace{2cm}
\begin{minipage}{\dimexpr0.33\linewidth-2\fboxsep-2\fboxrule\relax}
        \textsuperscript{2}Lister Hill National Center for Biomedical Communications, National Library of Medicine
\end{minipage}\\
National Institutes of Health, Bethesda, 20892-1182, USA\\
{\tt\small \{hoochang.shin; le.lu; kirk.roberts; rms\}@nih.gov; ddemner@mail.nih.gov; JYao@cc.nih.gov}
}

\maketitle

\begin{abstract}
   Despite the recent advances in automatically describing image contents, their applications have been mostly limited to image caption datasets containing natural images (e.g., Flickr 30k, MSCOCO).
   In this paper, we present a deep learning model to efficiently detect a disease from an image and annotate its contexts (e.g., location, severity and the affected organs).
   We employ a publicly available radiology dataset of chest x-rays and their reports, and use its image annotations to mine disease names to train convolutional neural networks (CNNs).
   In doing so, we adopt various regularization techniques to circumvent the large normal-vs-diseased cases bias.
   Recurrent neural networks (RNNs) are then trained to describe the contexts of a detected disease, based on the deep CNN features.
   Moreover, we introduce a novel approach to use the weights of the already trained pair of CNN/RNN on the domain-specific image/text dataset, to infer the joint image/text contexts for composite image labeling.
   Significantly improved image annotation results are demonstrated using the recurrent neural cascade model by taking the joint image/text contexts into account.
\end{abstract}

\begin{figure}[t]
\begin{center}
   \includegraphics[width=1.0\linewidth]{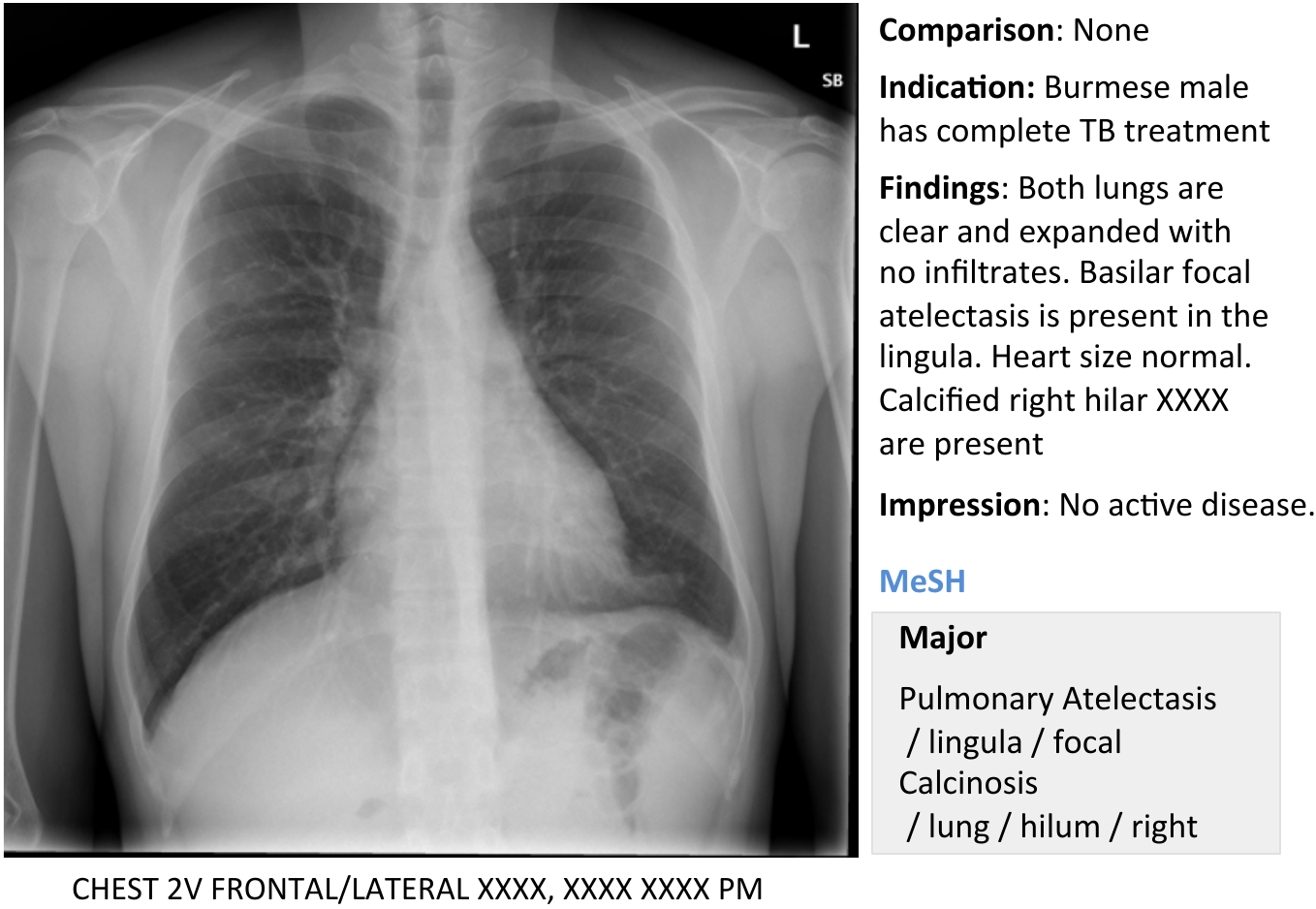}
\end{center}
   \caption{An example of OpenI \cite{openi} chest x-ray image, report, and annotations.}
\label{fig:chestx_example_01}
\end{figure}

\section{Introduction}

Comprehensive image understanding requires more than single object classification.
There have been many advances in automatic generation of image captions to describe image contents, which is closer to a more complete image understanding than classifying an image to a single object class. 
Our work is inspired by many of the recent progresses in image caption generation \cite{mao2015explain,socher2014grounded,kiros2015unifying,donahue2015long,vinyals2015show,fang2015captions,Chen_2015_CVPR,xu2015show,Karpathy_2015_CVPR}, as well as some of the earlier pioneering work \cite{kulkarni2013babytalk,feng2010many,farhadi2010every}.
The former have substantially improved performance, largely due to the introduction of ImageNet database \cite{deng2009imagenet} and to advances in deep convolutional neural networks (CNNs), effectively learning to recognize the images with a large pool of hierarchical representations.
Most recent work also adapt recurrent neural networks (RNNs), using the rich deep CNN features to generate image captions.
However, the applications of the previous studies were limited to natural image caption datasets such as Flickr8k \cite{hodosh2013framing}, Flickr30k \cite{young2014image}, or MSCOCO \cite{lin2014microsoft} which can be generalized from ImageNet.

Likewise, there have been continuous efforts and progresses in the automatic recognition and localization of specific diseases and organs, mostly on datasets where target objects are explicitly annotated \cite{shi2014joint,Hofmanninger_2015_CVPR,subbanna2014iterative,ngo2014fully,ledig2014patch,Rupprecht_2015_CVPR}.
Yet, learning from medical image text reports and generating annotations that describe diseases and their contexts have been very limited.
Nonetheless, providing a description of a medical image's content similar to a radiologist would describe could have a great impact.
A person can better understand a disease in an image if it is presented with its context, e.g., where the disease is, how severe it is, and which organ is affected.
Furthermore, a large collection of medical images can be automatically annotated with the disease context and the images can be retrieved based on their context, with queries such as ``\textit{find me images with pulmonary disease in the upper right lobe}''.

In this work, we demonstrate how to automatically annotate chest x-rays with diseases along with describing the contexts of a disease, e.g., location, severity, and the affected organs.
A publicly available radiology dataset is exploited which contains chest x-ray images and reports published on the Web as a part of the OpenI \cite{openi} open source literature and biomedical image collections.
An example of a chest x-ray image, report, and annotations available on OpenI is shown in Figure \ref{fig:chestx_example_01}.

A common challenge in medical image analysis is the data bias.
When considering the whole population, diseased cases are much rarer than healthy cases, which is also the case in the chest x-ray dataset used.
Normal cases account for 37\% (2,696 images) of the entire dataset (7,284 images), compared to the most frequent disease\footnote{Clinical \textit{findings}, \textit{disorders}, and other \textit{abnormal artifacts} will be collectively referred to as \textit{diseases} in this paper.} case ``\textit{opacity}'' which accounts for 12\% (840 images) and the next frequent ``\textit{cardiomegaly}'' constituting for 9\% (655 images).
In order to circumvent the normal-vs-diseased cases bias, we adopt various regularization techniques in CNN training.

In analogy to the previous works using ImageNet-trained CNN features for image encoding and RNNs to generate image captions, we first train CNN models with one disease label per chest x-ray inferred from image annotations, e.g., ``\textit{calcified granuloma}'', or ``\textit{cardiomegaly}''.
However, such single disease labels do not fully account for the context of a disease.
For instance, ``\textit{calcified granuloma in right upper lobe}'' would be labeled the same as the ``\textit{small calcified granuloma in left lung base}'' or ``\textit{multiple calcified granuloma}''.

Inspired by the ideas introduced in \cite{irsoy2014opinion,yao2015describing,ioffe2015batch,xu2015show,venugopalan2015sequence}, we employ the already trained RNNs to obtain the \textit{context} of annotations, and recurrently use this to infer the image labels with contexts as attributes.
Then we re-train CNNs with the obtained joint image/text contexts and generate annotations based on the new CNN features.
With this recurrent cascade model, image/text contexts are taken into account for CNN training (images with ``\textit{calcified granuloma in right upper lobe}'' and ``\textit{small calcified granuloma in left lung base}'' will be assigned different labels), to ultimately generate better and more accurate image annotations.

\section{Related Work}
This work was initially inspired by the early work in image caption generation \cite{kulkarni2013babytalk,feng2010many,farhadi2010every}, where we take more recently introduced ideas of using CNNs and RNNs \cite{mao2015explain,socher2014grounded,kiros2015unifying,donahue2015long,vinyals2015show,fang2015captions,Chen_2015_CVPR,xu2015show,Karpathy_2015_CVPR} to combine recent advances in computer vision and machine translation.
We also exploit the concepts of leveraging the mid-level RNN representations to infer image labels from the annotations \cite{irsoy2014opinion,yao2015describing,ioffe2015batch,xu2015show,venugopalan2015sequence}.

Methods for mining and predicting labels from radiology images and reports were investigated in \cite{shin2015interleaved,shin2015interleavedarx,syeda2015learning}.
However, the image labels were mostly limited to disease names and did not contain much contextual information.
Furthermore, the majority of cases in the datasets were diseased cases.
In reality, most cases are normal, so that it is a challenge to detect relatively rarer diseased cases within such unbalanced data.

Mining images and image labels from a large collections of photo streams and blog posts on the Web were demonstrated in \cite{Kim_2015_CVPR2,Kim_2015_CVPR1,kim2014joint} where images could be searched with natural language queries.
Associating neural word embeddings and deep image representations were explored in \cite{klein2015associating}, but generating descriptions from such images/text pairs or image/word embeddings have not yet been demonstrated.

Detecting diseases from x-rays was demonstrated in \cite{avni2011x,melendez2015novel,jaeger2014automatic}, classifying chest x-ray image views in \cite{xue2015chest}, and segmenting body parts in chest x-rays and computed tomography in \cite{Boussaid_2014_CVPR,hermann2014evaluation}.
However, learning image contexts from text and re-generating image descriptions similar to what a human would describe have not yet been studied.
To the best of our knowledge, this is the first study mining from a radiology image and report dataset, not only to classify and detect images but also to describe their context.

\section{Dataset}

We use a publicly available radiology dataset of chest x-rays and reports that is a subset of the OpenI \cite{openi} open source literature and biomedical image collections.
It contains 3,955 radiology reports from the Indiana Network for Patient Care, and 7,470 associated chest x-rays from the hospitals' picture archiving systems.
The entire dataset has been fully anonymized via an aggressive anonymization scheme, which achieved 100\% precision in de-identification. However, a few findings have been rendered uninterpretable.
More details about the dataset and the anonymization procedure can be found in \cite{demner2015preparing}, and an example case of the dataset is shown in Figure \ref{fig:chestx_example_01}.

Each report is structured as \textit{comparison}, \textit{indication}, \textit{findings}, and \textit{impression} sections, in line with a common radiology reporting format for diagnostic chest x-rays.
In the example shown in Figure \ref{fig:chestx_example_01}, we observe an error resulting from the aggressive automated de-identification scheme.
A word possibly indicating a disease was falsely detected as a personal information, and was thereby ``anonymized'' as ``XXXX''.
While radiology reports contain comprehensive information about the image and the patient, they may also contain information that cannot be inferred from the image content.
For instance, in the example shown in Figure \ref{fig:chestx_example_01}, it is probably impossible to determine that the image is of a Burmese male.

On the other hand, a manual annotation of MEDLINE \textsuperscript{\textregistered} citations with controlled vocabulary terms (Medical Subject Headings (MeSH\textsuperscript{\textregistered}) \cite{mesh}) is known to significantly improve the quality of the image retrieval results \cite{haynes1994developing,hersh2009trec,darmoni2012improving}.
MeSH terms for each radiology report in OpenI (available for public use) are annotated according to the process described in \cite{demnerannotation}.
We use these to train our model.

Nonetheless, it is impossible to assign a single image label based on MeSH and train a CNN to reproduce them, because MeSH terms seldom appear individually when describing an image.
The twenty most frequent MeSH terms appear with other terms in more than 60\% of the cases.
Normal cases (term ``\textit{normal}'') on the contrary, do not have any overlap, and account for 37\% of the entire dataset.
The thirteen most frequent MeSH terms appearing more than 180 times are provided in Table \ref{tab:meshterms}, along with the total number of cases in which they appear, the number of cases they overlap with in an image and the overlap percentages.
The x-ray images are provided in Portable Network Graphics (PNG) format, with sizes varying from $512\times 420$ to $512\times 624$.
We rescale all CNN input training and testing images to a size of $256\times 256$.

\begin{table}
\begin{center}
\resizebox{1\linewidth}{!}{
\begin{tabular}{|l||c|c|c|}
\hline
MeSH Term                       & Total & Overlap & Overlap Percent \\
\hline\hline
normal                          & 2,696 & 0       & 0\% \\
\hline
opacity                         & 840   & 666     & 79\% \\
\hline
cardiomegaly                    & 655   & 492     & 75\% \\
\hline
calcinosis                      & 558   & 444     & 80\% \\
\hline
lung/hypoinflation              & 539   & 361     & 67\% \\
\hline
calcified granuloma             & 511   & 303     & 59\% \\
\hline
thoracic vertebrae/degenerative & 471   & 296     & 63\% \\
\hline
lung/hyperdistention            & 400   & 260     & 65\% \\
\hline
spine/degenerative              & 337   & 219     & 65\% \\
\hline
catheters, indwelling           & 222   & 159     & 72\% \\
\hline
granulomatous disease           & 213   & 165     & 78\% \\
\hline
nodule                          & 211   & 160     & 76\% \\
\hline
surgical instruments            & 180   & 120     & 67\% \\
\hline
\end{tabular}}
\end{center}
\caption{Thirteen most frequent MeSH terms appearing over 180 times, and the number of the terms mentioned with other terms (overlap) in an image and their percentages.}
\label{tab:meshterms}
\end{table}

\section{Disease Label Mining}

The CNN-RNN based image caption generation approaches \cite{mao2015explain,socher2014grounded,kiros2015unifying,donahue2015long,vinyals2015show,fang2015captions,Chen_2015_CVPR,xu2015show,Karpathy_2015_CVPR} require a well-trained CNN to encode input images effectively.
Unlike natural images that can simply be encoded by ImageNet-trained CNNs, chest x-rays differ significantly from the ImageNet images.
In order to train CNNs with chest x-ray images, we sample some frequent annotation patterns with less overlaps for each image, in order to assign image labels to each chest x-ray image and train with cross-entropy criteria.
This is similar to the previous works from \cite{shin2015interleaved,shin2015interleavedarx,syeda2015learning}, which mines disease labels of images from their annotation text (radiology reports).


We find 17 unique patterns of MeSH term combinations appearing in 30 or more cases.
This allows us to split the dataset in training/validation/testing cases as 80\%/10\%/10\% and place at least 10 cases each in the validation and testing sets.
They include the terms shown in Table \ref{tab:meshterms}, as well as \textit{scoliosis}, \textit{osteophyte}, \textit{spondylosis}, \textit{fractures/bone}.
MeSH terms appearing frequently but without unique appearance patterns include \textit{pulmonary atelectasis}, \textit{aorta/tortuous}, \textit{pleural effusion}, \textit{cicatrix}, etc.
They often appear with other disease terms (e.g. \textit{consolidation}, \textit{airspace disease}, \textit{atherosclerosis}).
We retain about 40\% of the full dataset with this disease image label mining, where the annotations for the remaining 60\% of images are more complex (and it is therefore difficult to assign a single disease label).

\section{Image Classification with CNN}
\label{sec:cnn_training}

We use the aforementioned 17 unique disease annotation patterns (in Table \ref{tab:meshterms}, and \textit{scoliosis}, \textit{osteophyte}, \textit{spondylosis}, \textit{fractures/bone}) to label the images and train CNNs.
For this purpose, we adopt various regularization techniques to deal with the normal-vs-diseased cases bias.
For our default CNN model we chose the simple yet effective Network-In-Network (NIN) \cite{lin2013network} model because the model is small in size, fast to train, and achieves similar or better performance to the most commonly used AlexNet model \cite{krizhevsky2012imagenet}.
We then test whether our data can benefit from a more complex state-of-the-art CNN model, i.e. GoogLeNet \cite{Szegedy_2015_CVPR}.

From the 17 chosen disease annotation patterns, normal cases account for 71\% of all images, well above the numbers of cases for the remaining 16 disease annotation patterns.
We balance the number of samples for each case by augmenting the training images of the smaller cases where we randomly crop $224\times 224$ size images from the original $256\times 256$ size image.

\subsection{Regularization by Batch Normalization and Data Dropout}

Even when we balance the dataset by augmenting many diseased samples, it is difficult for a CNN to learn a good model to distinguish many diseased cases from normal cases which have many variations on their original samples.
It was shown in \cite{ioffe2015batch} that normalizing via mini-batch statistics during training can serve as an effective regularization technique to improve the performance of a CNN model.
By normalizing via mini-batch statistics, the training network was shown not to produce deterministic values for a given training example, thereby regularizing the model to generalize better.

Inspired by this and by the concept of Dropout \cite{hinton2012improving}, we regularize the normal-vs-diseased cases bias via randomly dropping out an excessive proportion of normal cases compared to the frequent diseased pattern when sampling mini-batches.
We then normalize according to the mini-batch statistics where each mini-batch consists of a balanced number of samples per disease case and a random selection of normal case samples.
The number of samples for disease cases is balanced by random cropping during training, where each image of a diseased case is augmented at least four times.

We test both regularization techniques to assess their effectiveness on our dataset.
The training and validation accuracies of the NIN model with batch-normalization, data-dropout, and both are provided in Table \ref{tab:acc_nin_regularization}.
While batch-normalization and data-dropout alone do not significantly improve performance, combining both increases the validation accuracy by about 2\%.

\begin{table}
\begin{center}
\resizebox{1.0\linewidth}{!}{
\begin{tabular}{|l||c|c|}
\hline
                                  & training accuracy & validation accuracy\\
\hline\hline
NIN with batch-normalization (BN) &   94.06\%         & 56.65\%            \\
\hline     
NIN with data-dropout (DDropout)  &   98.78\%         & 58.38\%            \\
\hline
NIN with BN and DDropout          &   100.0\%         & 62.43\%            \\
\hline
\end{tabular}}
\end{center}
\caption{Training and validation accuracy of NIN model with batch-normalization, data-dropout, and both. Diseased cases are very limited compared to normal cases, leading to overfitting even with regularizations.}
\label{tab:acc_nin_regularization}
\end{table}

\subsection{Effect of Model Complexity}
\label{sec:cnn_model_complexity}

\begin{table}
\begin{center}
\resizebox{1.0\linewidth}{!}{
\begin{tabular}{|l||c|c|}
\hline
                                  & training accuracy & validation accuracy\\
\hline\hline
GoogLeNet with BN and DDropout    &     98.11\%         & 66.40\%            \\
\hline
GoogLeNet with BN, DDropout, No-Crop  &   100.0\%         & 69.84\%            \\
\hline
\end{tabular}}
\end{center}
\caption{Training and validation accuracy of GoogLeNet model with batch-normalization, data-dropout, and without cropping the images for data augmentation.}
\label{tab:acc_model_complexity}
\end{table}

We also validate whether the dataset can benefit from a more complex GoogLeNet \cite{Szegedy_2015_CVPR}, which is arguably the current state-of-the-art CNN architecture.
We apply both batch-normalization and data-dropout, and follow recommendations suggested in \cite{ioffe2015batch} (where human accuracy on the ImageNet dataset is superseded): increase learning rate, remove dropout, remove local response normalization.
The final training and validation accuracies using GoogLeNet model are provided in Table \ref{tab:acc_model_complexity}, where we achieve a higher ($\sim 4$\%) accuracy\footnote{The final models are trained with default learning rate of $1.0$, with \textit{step down} learning rate scheduling decreasing the learning rate by 50\% and 33\% each for NIN and GoogLeNet model in every 1/3th of the total $100$ training epochs.
We could not achieve high enough validation accuracy using exponential learning rate decay as in \cite{ioffe2015batch}.}.
We also observe a further $\sim 3$\% increase in accuracy when the images are no longer cropped, but merely duplicated to balance the dataset.

\section{Annotation Generation with RNN}
\label{sec:annotation_rnn}

\begin{figure}[t]
\begin{center}
   \includegraphics[width=1.0\linewidth]{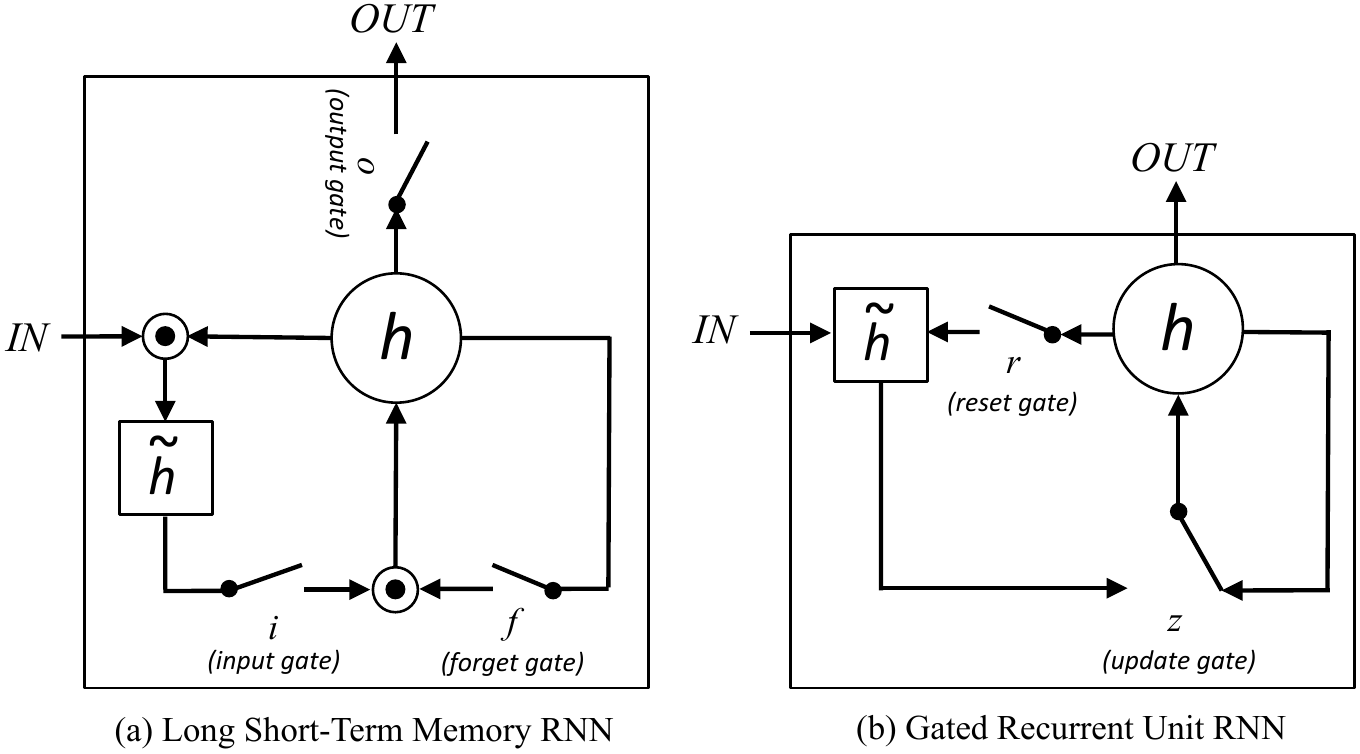}
\end{center}
   \caption{Simplified illustrations of (a) Long Short-Term Memory (LSTM) and (b) Gated Recurrent Unit (GRU) RNNs. The illustrations are adapted and modified from Figure 2 in \cite{chung2014empirical}.}
\label{fig:lstm_rnn_illustrations}
\end{figure}

We use recurrent neural networks (RNNs) to learn the annotation sequence given input image CNN embeddings.
We test both Long Short-Term Memory (LSTM) \cite{hochreiter1997long} and Gated Recurrent Unit (GRU) \cite{cho2014learning} implementations of RNNs.
Simplified illustrations of LSTM and GRU are shown in Figure \ref{fig:lstm_rnn_illustrations}, and the details of both RNN implementations are briefly introduced below.

\subsection{Recurrent Neural Network Implementations}

\paragraph{Long Short-Term Memory}

The LSTM implementation, originally proposed in \cite{hochreiter1997long}, has been successfully applied to speech recognition \cite{graves2013speech}, sequence generation  \cite{graves2013generating}, machine translation \cite{cho2014learning,sutskever2014sequence,luong2015addressing}, and several image caption generation works mentioned in the main paper.
LSTM is defined by the following equations:

\begin{align}
  i_t &= \sigma(W^{(i)}x_t + U^{(i)}m_{t-1})\\
  f_t &= \sigma(W^{(f)}x_t + U^{(f)}m_{t-1})\\
  o_t &= \sigma(W^{(o)}x_t + U^{(o)}m_{t-1})\\
  \widetilde{h}_t &= \text{tanh}(W^{(h)}x_t + U^{(h)}m_{t-1})\\
  h_t &= f_t \odot h_{t-1} + i_t \odot \widetilde{h}_t\\
  m_t &= o_t \odot \text{tanh}(h_t)
\label{eq:lstm}
\end{align}

\noindent
where $i_t$ is the input gate, $f_t$ the forget gate, $o_t$ the output gate, $h_t$ the state vector (memory), $\widetilde{h}_t$ the new state vector (new memory), and $m_t$ the output vector.
$W$ is a matrix of trained parameters (weights), and $\sigma$ is the logistic sigmoid function.
$\odot$ represents the product of a vector with a gate value.

Please note that the notation used for the memory ($h_t, \widetilde{h}_t$) and output ($m_t$) vectors differs from that in \cite{hochreiter1997long} and the other previous work.
Our notation is intended to simplify the annotations and figures comparing LSTM to GRU.

\paragraph{Gated Recurrent Unit}

The GRU implementation has been proposed most recently in \cite{cho2014learning}, where it was successfully applied to machine translation.
GRU is defined by the following equations:

\begin{align}
  z_t &= \sigma(W^{(z)}x_t + U^{(z)}h_{t-1})\\
  r_t &= \sigma(W^{(r)}x_t + U^{(r)}h_{t-1})\\
  \widetilde{h}_t &= \text{tanh}(Wx_t + r_t \odot Uh_{t-1})\\
  h_t &= z_t \odot h_{t-1} + (1-z_t) \odot \widetilde{h}_t
\end{align}

\noindent
where $z_t$ is the update gate, $r_t$ the reset gate, $\widetilde{h}_t$ the new state vector, and $h_t$ the final state vector.

\subsection{Training}

The number of MeSH terms describing diseases ranges from 1 to 8 (except \textit{normal} which is one word), with a mean of 2.56 and standard deviation of 1.36.
The majority of descriptions contain up to five words.
Since only 9 cases have images with descriptions longer than 6 words, we ignore these by constraining the RNNs to unroll up to 5 time steps.
We zero-pad annotations with less than five words with the end-of-sentence token to fill in the five word space.

The parameters of the gates in LSTM and GRU decide whether to update their current state $\mathbf{h}$ to the new candidate state $\widetilde{\mathbf{h}}$, where these are learned from the previous input sequences.
Further details about LSTM can be found in \cite{hochreiter1997long,fang2015captions,donahue2015long,vinyals2015show}, and about GRU and its comparisons to LSTM in \cite{cho2014learning,jozefowicz2015empirical,chung2015gated,chung2014empirical,karpathy2015visualizing}.
We set the initial state of RNNs as the CNN image embedding (CNN($I$)), and the first annotation word as the initial input. 
The output of the RNNs are the following annotation word sequences, and we train RNNs by minimizing the negative log likelihood of output sequences and true sequences:

\begin{equation}
  L(I,S) = -\sum_{t=1}^{N} \{P_\text{RNN}(y_t=s_t) | \text{CNN}(I)\},
\label{eq:rnn_training}
\end{equation}

\noindent
where $y_t$ is the output word of RNN in time step $t$, $s_t$ the correct word, $\text{CNN(I)}$ the CNN embedding of input image $I$, and $N$ the number of words in the annotation ($N=5$ with the end-of-sequence zero-padding).
Equation \ref{eq:rnn_training} is not a true conditional probability (because we only initialize the RNNs' state vector to be CNN($I$)) but a convenient way to describe the training procedure.

Unlike the previous work \cite{Karpathy_2015_CVPR,fang2015captions,donahue2015long} where they use the last (FC-8) or second last (FC-7) fully-connected layers of AlexNet \cite{krizhevsky2012imagenet} or VGG-Net \cite{simonyan2014very} model, the NIN or GoogLeNet models replace the fully-connected layers with the average-pooling layers \cite{lin2013network,Szegedy_2015_CVPR}.
We therefore use the output of the last spatial average-pooling layer as the image embedding to initialize the RNN state vectors.
The size of our RNNs' state vectors are $\mathbb{R}^{1\times 1024}$, which is identical to the output size of the average-pooling layers from NIN and GoogLeNet.

\subsection{Sampling}
\label{sec:sampling}

In sampling we again initialize the RNN state vectors with the CNN image embedding ($\mathbf{h}_{t=1}$=CNN($I$)).
We then use the CNN prediction of the input image as the first word as the input to the RNN, to sample following sequences up to five words.
As previously, images are normalized by the batch statistics before being fed to the CNN.
We use GoogLeNet as our default CNN model since it performs better than the NIN model in Sec. \ref{sec:cnn_model_complexity}.

\subsection{Evaluation}

\begin{table}
\begin{center}
\resizebox{1.0\linewidth}{!}{
\begin{tabular}{|l||c|c|c|c|c|}
\hline
               & train                   & validation              & test \\ 
               \cline{2-4}
               & BLEU -1/ -2/ -3 / -4    & BLEU -1/ -2/ -3 / -4    & BLEU -1/ -2/ -3 / -4 \\
\hline\hline
LSTM           & 82.6 / 19.2 / 2.2 / 0.0 & 67.4 / 15.1 / 1.6 / 0.0 & 78.3 / 0.4 / 0.0 / 0.0  \\
\hline
GRU            & 98.9 / 46.9 / 1.2 / 0.0 & 85.8 / 14.1 / 0.0 / 0.0 & 75.4 / 3.0 / 0.0 / 0.0 \\
\hline
\end{tabular}}
\end{center}
\caption{BLEU scores validated on the training, validation, test set, using LSTM and GRU RNN models for the sequence generation.}
\label{tab:bleu_scores_iter0}
\end{table}

We evaluate the annotation generation on the BLEU \cite{papineni2002bleu} score averaged over all of the images and their annotations in the training, validation, and test set.
BLEU scores is a metric measuring a modified form of precision to compare n-gram words of generated and reference sentences.
The BLEU scores evaluated are provided in Table \ref{tab:bleu_scores_iter0}.
The BLEU-$N$ scores are evaluated for cases with $\geq N$ words in the annotations, using the implementation of \cite{bird2009natural}.

We noticed that LSTM is easier to train, while GRU model yields better results with more carefully selected hyper-parameters\footnote{The final LSTM models are obtained with -- learning rate: $2\times 10^{-3}$, learning rate decay: 0.97, decay rate: 0.95, without dropout; and the final GRU model is obtained with -- learning rate: $1\times 10^{-4}$, learning rate decay: 0.99, decay rate: 0.99, with dropout rate: 0.9.
With the same setting, adding dropout to LSTM model has adverse effect on its validation loss, similarly when increasing the number of LSTM layers to 3.
The number of layers are 2 for both RNN models, and they are both trained with the batch size of 50.}.
While we find it difficult to conclude which model is better, the GRU model seems to achieve higher scores on average.

\begin{figure*}[t]
\begin{center}
   \includegraphics[width=.9\linewidth]{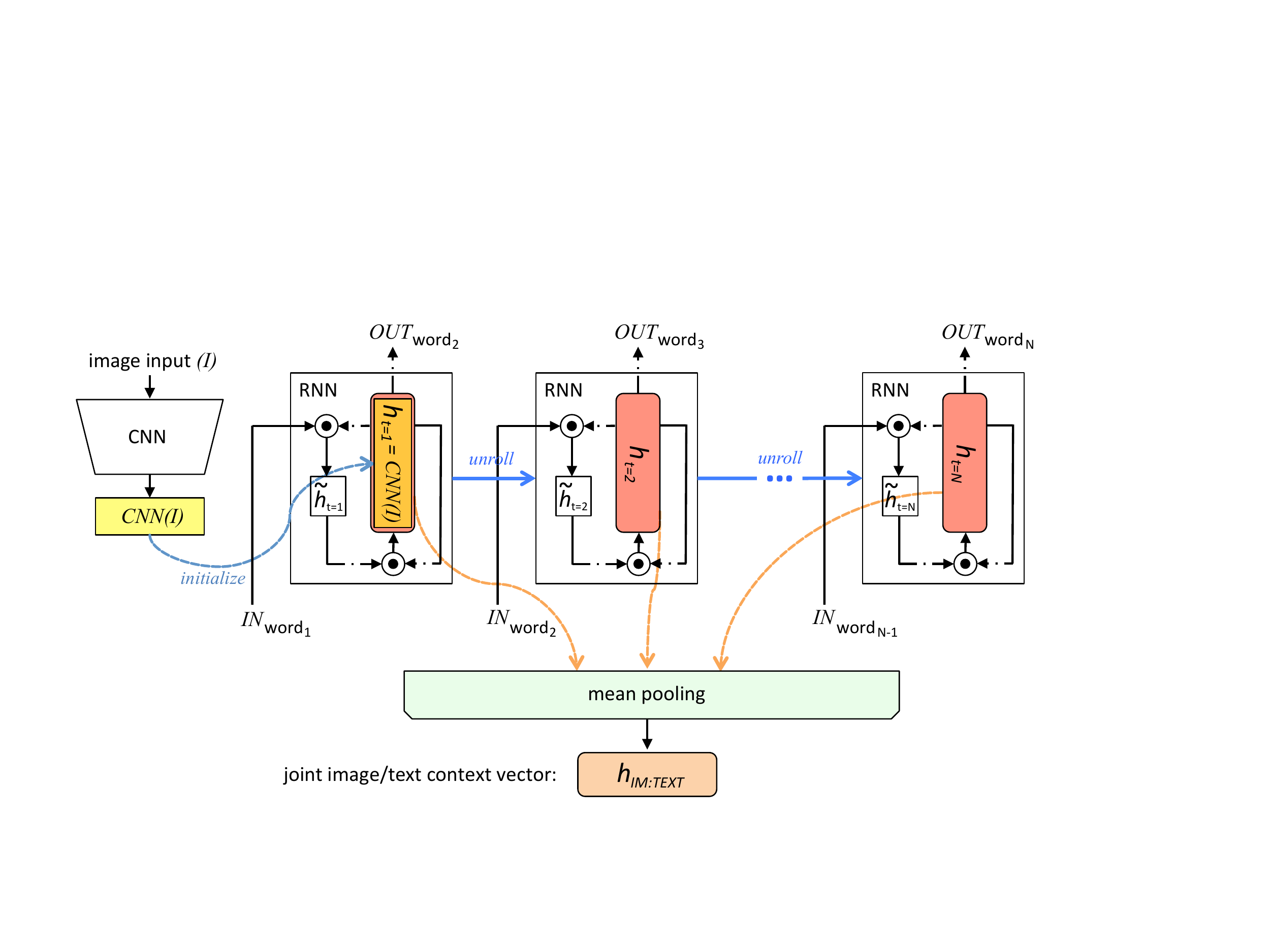}
\end{center}
   \caption{An illustration of how joint image/text context vector is obtained. RNN's state vector ($h$) is initialized with the CNN image embedding (CNN($I$)), and it's unrolled over the annotation sequences with the words as input. Mean-pooling is applied over the state vectors in each word of the sequence, to obtain the joint image/text context vector. All RNNs share the same parameters, which are trained in the first round.}
\label{fig:cnn_rnn_for_context}
\end{figure*}

\section{Recurrent Cascade Model for Image Labeling with Joint Image/Text Context}

In Section \ref{sec:cnn_training}, our CNN models are trained with disease labels only where the context of diseases are not considered.
For instance, the same \textit{calcified granuloma} label is assigned to all image cases that actually may describe the disease differently in a finer semantic level, such as ``\textit{calcified granuloma in right upper lobe}'', ``\textit{small calcified granuloma in left lung base}'', and ``\textit{multiple calcified granuloma}''.

Meanwhile, the RNNs trained in Section \ref{sec:annotation_rnn} encode the text annotation sequences given the CNN embedding of the image the annotation is describing.
We therefore use the already trained CNN and RNN to infer better image labels, integrating the contexts of the image annotations beyond just the name of the disease.
This is achieved by generating \textit{joint image/text context vectors} that are computed by applying mean-pooling on the state vectors ($h$) of RNN at each step over the annotation sequence.
Note, that the state vector of RNN is initialized with the CNN image embeddings (CNN($I$)), and the RNN is unrolled over the annotation sequence, taking each word of the annotation as input.
The procedure is illustrated in Figure \ref{fig:cnn_rnn_for_context}, and the RNNs share the same parameters.

\begin{figure}[t]
\begin{center}
   \includegraphics[width=1.0\linewidth]{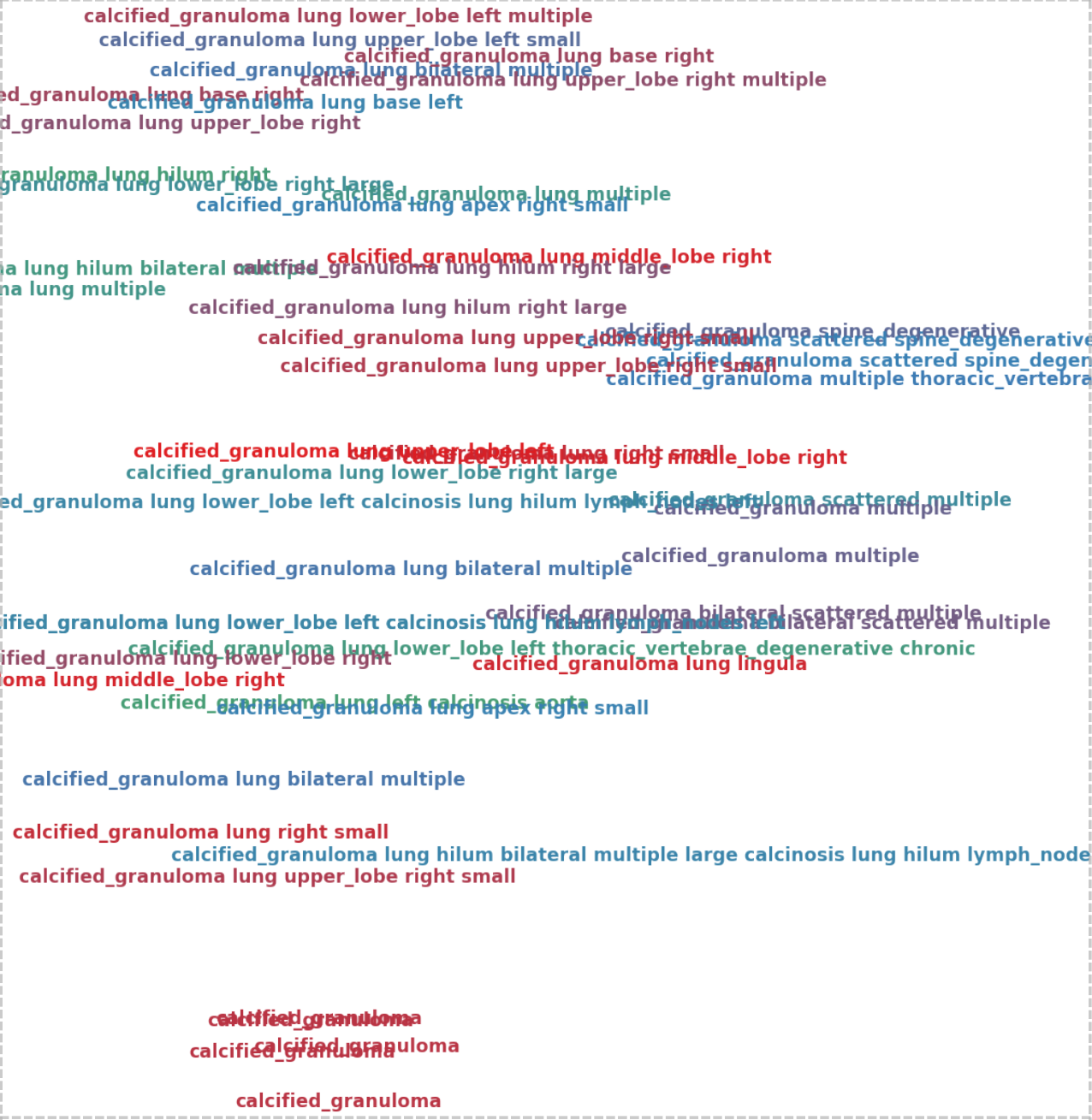}
\end{center}
   \caption{Visualization of joint image/text context vectors of about 50 samples of the annotations describing disease \textit{calcified granuloma} on 2D planes. Dimension reduction ($\mathbb{R}^{1\times 1024} \rightarrow \mathbb{R}^{1\times 2}$) is performed using t-SNE \cite{van2008visualizing}. Annotations describing a same disease can be divided into different ``labels'' based on their joint image/text contexts.}
\label{fig:imtext_context_vector_viz}
\end{figure}

The obtained \textit{joint image/text context vector} ($h_\text{im:text}$) encodes the image context as well as the text context describing the image.
Using a notation similar to Equation \ref{eq:rnn_training}, the \textit{joint image/text context vector} can be written as:

\begin{equation}
  h_\text{im:text} = \frac{\sum_{t=1}^{N} \{h_\text{RNN}(x_t) | \text{CNN}(I)\}}{N},
\label{eq:image_text_joint_context_vector}
\end{equation}

\noindent
where $x_t$ is the input word in the annotation sequence with $N$ words.
Different annotations describing a disease are thereby separated into different categories by the $h_\text{im:text}$, as shown in Figure \ref{fig:imtext_context_vector_viz}.
In Figure \ref{fig:imtext_context_vector_viz}, the $h_\text{im:text}$ vectors of about fifty annotations describing \textit{calcified granuloma} are projected onto two-dimensional planes via dimensionality reduction ($\mathbb{R}^{1\times 1024} \rightarrow \mathbb{R}^{1\times 2}$), using the t-SNE \cite{van2008visualizing} implementation of \cite{scikit-learn}.
We use the GRU implementation of the RNN because it showed better overall BLEU scores in Table \ref{tab:bleu_scores_iter0}.
A visualization example for the annotations describing \textit{opacity} can be found in the supplementary material.

From the $h_\text{im:text}$ generated for each of the image/annotation pair in the training and validation sets, we obtain new image labels taking disease context into account.
In addition, we are no longer limited to disease annotation mostly describing a single disease.
The joint image/text context vector $h_\text{im:text}$ summarizes both the image's context and word sequence, so that annotations such as ``\textit{calcified granuloma in right upper lobe}'', ``\textit{small calcified granuloma in left lung base}'', and ``\textit{multiple calcified granuloma}'' have different vectors based on their contexts.

Additionally, disease labels used in Section \ref{sec:cnn_training} with unique annotation patterns now have more cases, as cases with a disease described by different annotation words are no longer filtered out.
For example, \textit{calcified granuloma} previously had only 139 cases because cases with multiple diseases mentioned or with long description sequences were filtered out.
At present, 414 cases are associated with \textit{calcified granuloma}.
Likewise, opacity now has 207 cases, as opposed to the previous 65.
The average number of cases all first-mentioned disease labels has is $83.89$, with a standard deviation of $86.07$, a maximum of $414$ (\textit{calcified granuloma}) and a minimum of $18$ (\textit{emphysema}).

\begin{figure}[t]
\begin{center}
\includegraphics[width=1\linewidth]{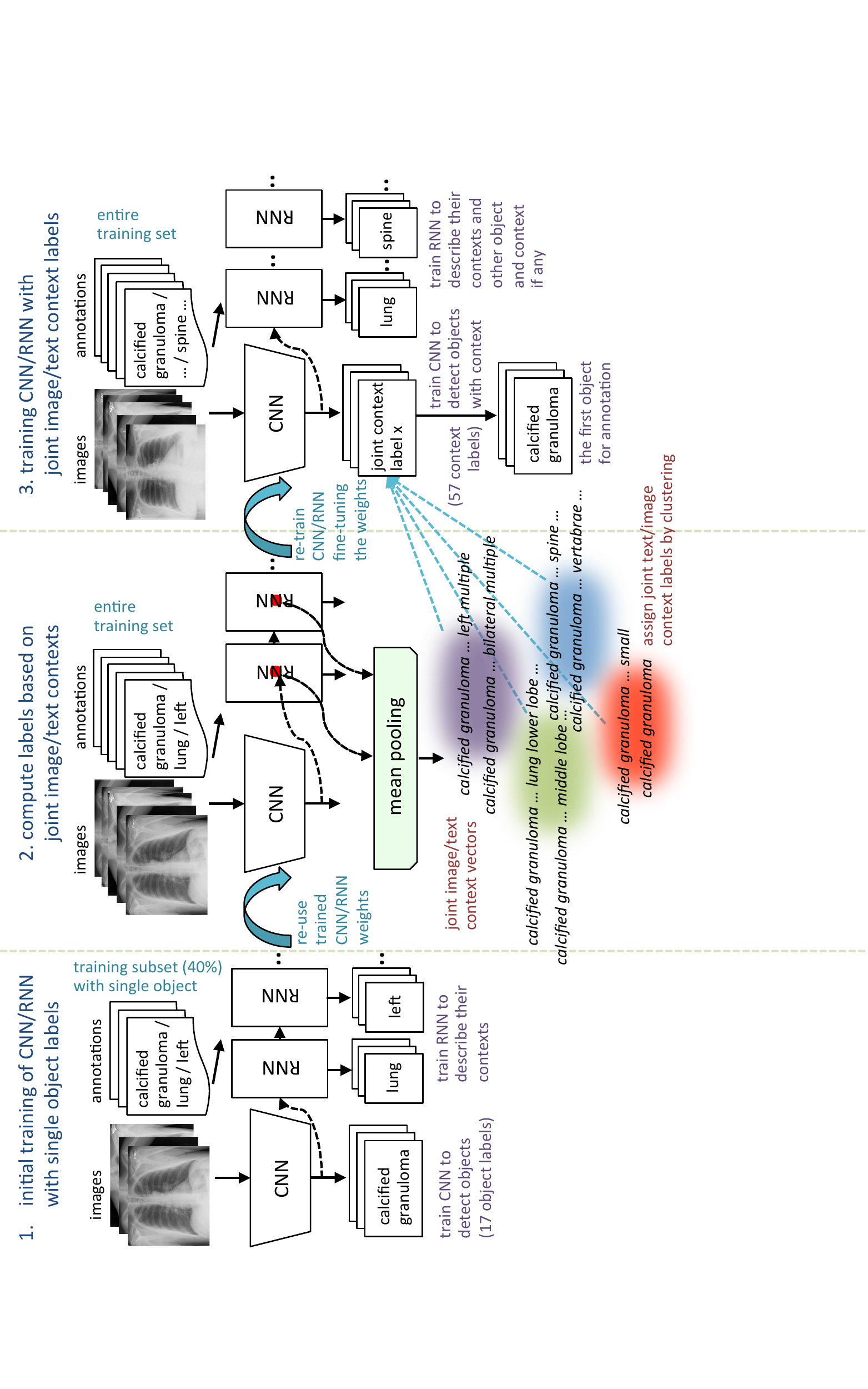}
\end{center}
   \caption{Overall workflow for the automated image annotation.}
\label{fig:workflow}
\end{figure}

For a disease label having more than 170 cases ($n\geq 170=$ (\textit{average+\textit{standard deviation}})), we divide the cases into sub-groups of more than 50 cases by applying \textit{k-means} clustering to the $h_\text{im:text}$ vector with $k=\text{Round}(n/50)$.
We train the CNN once more with the additional labels ($57$, compared to $17$ in Section \ref{sec:cnn_training}), train the RNN with the new CNN image embedding, and finally generate image annotations.
The new RNN training cost function (compared to Equation \ref{eq:rnn_training}) can be expressed as:

\begin{align}
  L&(I,S) = \nonumber\\
  &-\sum_{t=1}^{N} \left[ P_{\text{RNN}_{\text{iter}=1}}(y_t=s_t) \mid \{\text{CNN}_{\text{iter}=1}(I)|h_{\text{im:text}_{\text{iter}=0}} \}\right],
\label{eq:rnn_training_2}
\end{align}

\noindent
where $h_{\text{im:text}_{\text{iter}=0}}$ denotes the joint image/text context vector obtained from the first round (with limited cases and image labels at \textit{0th iteration}) of CNN and RNN training.
In the second CNN training round (\textit{1st iteration}), we fine-tune from the previous $\text{CNN}_{\text{iter}=0}$, by replacing the last classification layer with the new set of labels ($17 \rightarrow 57$) and training it with a lower learning rate ($0.1$), except for the classification layer.
The overall workflow is illustrated in Figure \ref{fig:workflow}.

\begin{figure*}[t]
\begin{center}
   \includegraphics[width=1.0\linewidth]{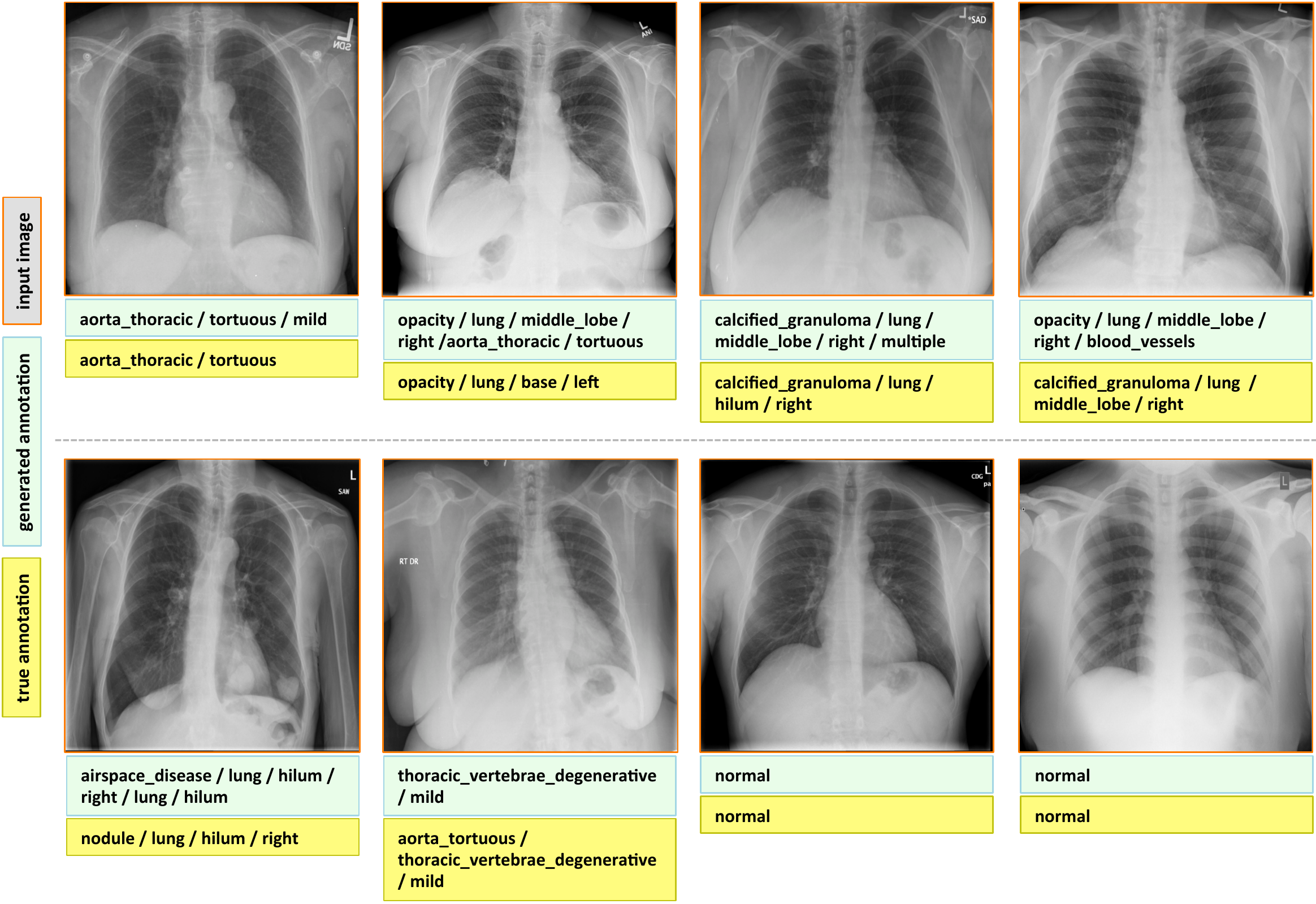}
\end{center}
   \caption{Examples of annotation generations (light green box) compared to true annotations (yellow box) for input images in the test set.}
\label{fig:annotation_generation_examples}
\end{figure*}

\subsection{Evaluation}

The final evaluated BLEU scores are provided in Table \ref{tab:bleu_scores_iter1}.
We achieve better overall BLEU scores than those in Table \ref{tab:bleu_scores_iter0} before using the joint image/text context.
It is noticeable that higher BLEU-N ($N>1$) scores are achieved compared to Table \ref{tab:bleu_scores_iter0}, indicating that more comprehensive image contexts are taken into account for the CNN/RNN training.
Also, slightly better BLEU scores are obtained using GRU on average and higher BLEU-1 scores are acquired using LSTM, although the comparison is empirical.
Examples of generated annotations on the chest x-ray images are shown in Figure \ref{fig:annotation_generation_examples}.
These are generated using the GRU model, and more examples can be found in the supplementary material.

\begin{table}
\begin{center}
\resizebox{1.0\linewidth}{!}{
\begin{tabular}{|l||c|c|c|c|c|}
\hline
               & train                     & validation                & test \\ 
               \cline{2-4}
               & BLEU -1/ -2/ -3 / -4      & BLEU -1/ -2/ -3 / -4      & BLEU -1/ -2/ -3 / -4 \\
\hline\hline
LSTM           & 97.2 / 67.1 / 14.9 / 2.8  & 68.1 / 30.1 / 5.2 / 1.1  & 79.3 / 9.1 / 0.0 / 0.0 \\
\hline
GRU            & 89.7 / 61.7 / 28.5 / 11.0 & 61.9 / 29.6 / 11.3 / 2.0 & 78.5 / 14.4 / 4.7 / 0.0  \\
\hline
\end{tabular}}
\end{center}
\caption{BLEU scores validated on the training, validation, test set, using LSTM and GRU RNN models trained on the first iteration, for the sequence generation.}
\label{tab:bleu_scores_iter1}
\end{table}

\section{Conclusion}

We present an effective framework to learn, detect disease, and describe their contexts from the patient chest x-rays and their accompanying radiology reports with Medical Subject Headings (MeSH) annotations.
Furthermore, we introduce an approach to mine joint contexts from a collection of images and their accompanying text, by summarizing the CNN/RNN outputs and their states on each of the image/text instances.
Higher performance on text generation is achieved on the test set if the joint image/text contexts are exploited to re-label the images and to train the proposed CNN/RNN framework subsequently.

To the best of our knowledge, this is the first study that mines from a publicly available radiology image and report dataset, not only to classify and detect disease in images but also to describe their context similar to a human observer would read.
While we only demonstrate on a medical dataset, the suggested approach could also be applied to other application scenario with datasets containing co-existing pairs of images and text annotations, where the domain-specific images differ from those of the ImageNet.

\section*{Acknowledgement}

This work was supported in part by the Intramural Research Program of the National Institutes of Health Clinical
Center, and in part by a grant from the KRIBB Research Initiative Program (Korean Biomedical Scientist Fellowship Program), Korea Research Institute of Bioscience and Biotechnology, Republic of Korea. We thank NVIDIA for the K40 GPU donation.

{\small
\bibliographystyle{ieee}
\bibliography{egbib}
}

\newpage

\begin{appendices}


\maketitle

\section{More Examples of the Chest X-ray Dataset}

\begin{figure}[h]
\begin{center}
   \includegraphics[width=1.0\linewidth]{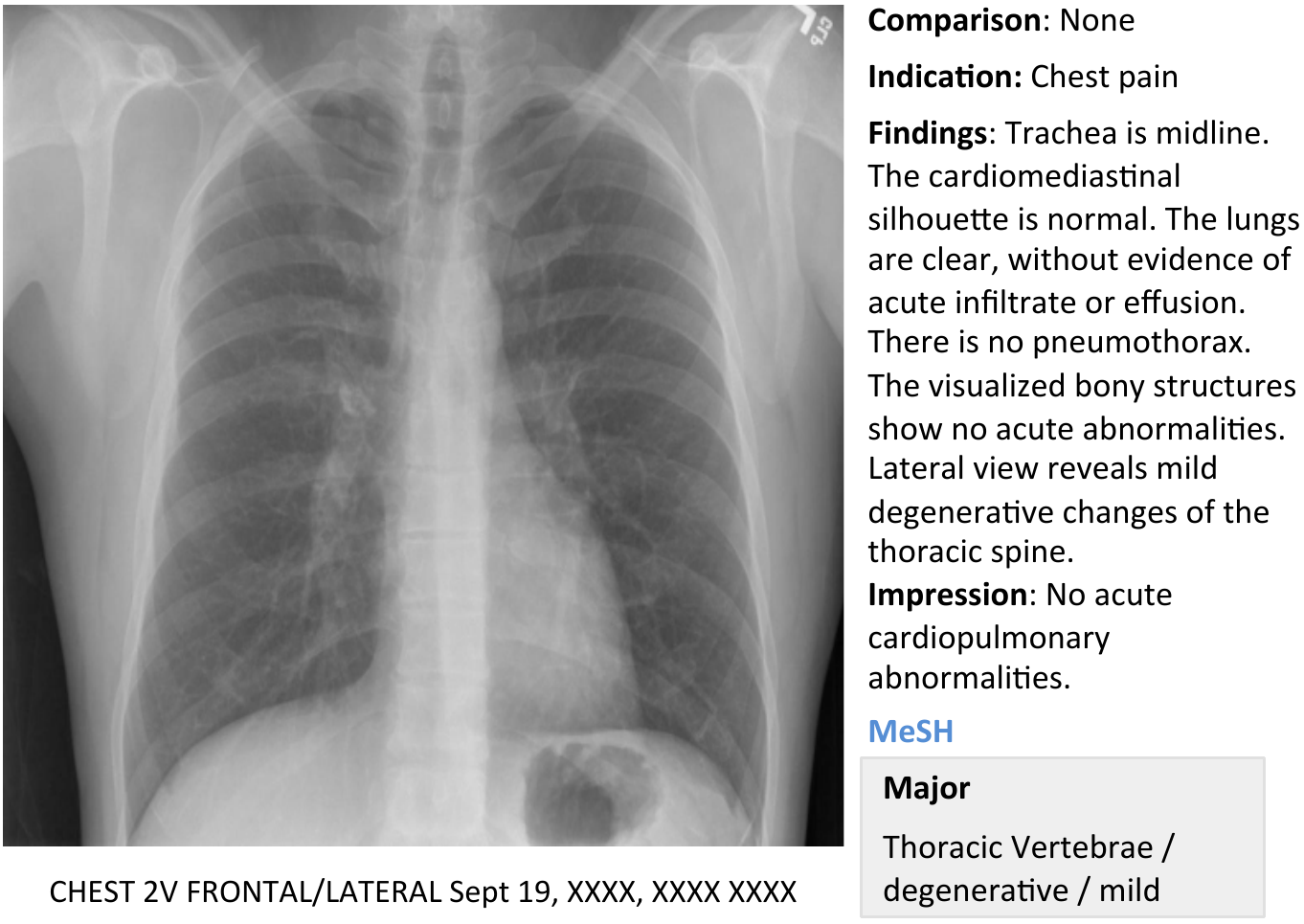}\\
   \vspace{.5cm}
   \includegraphics[width=1.0\linewidth]{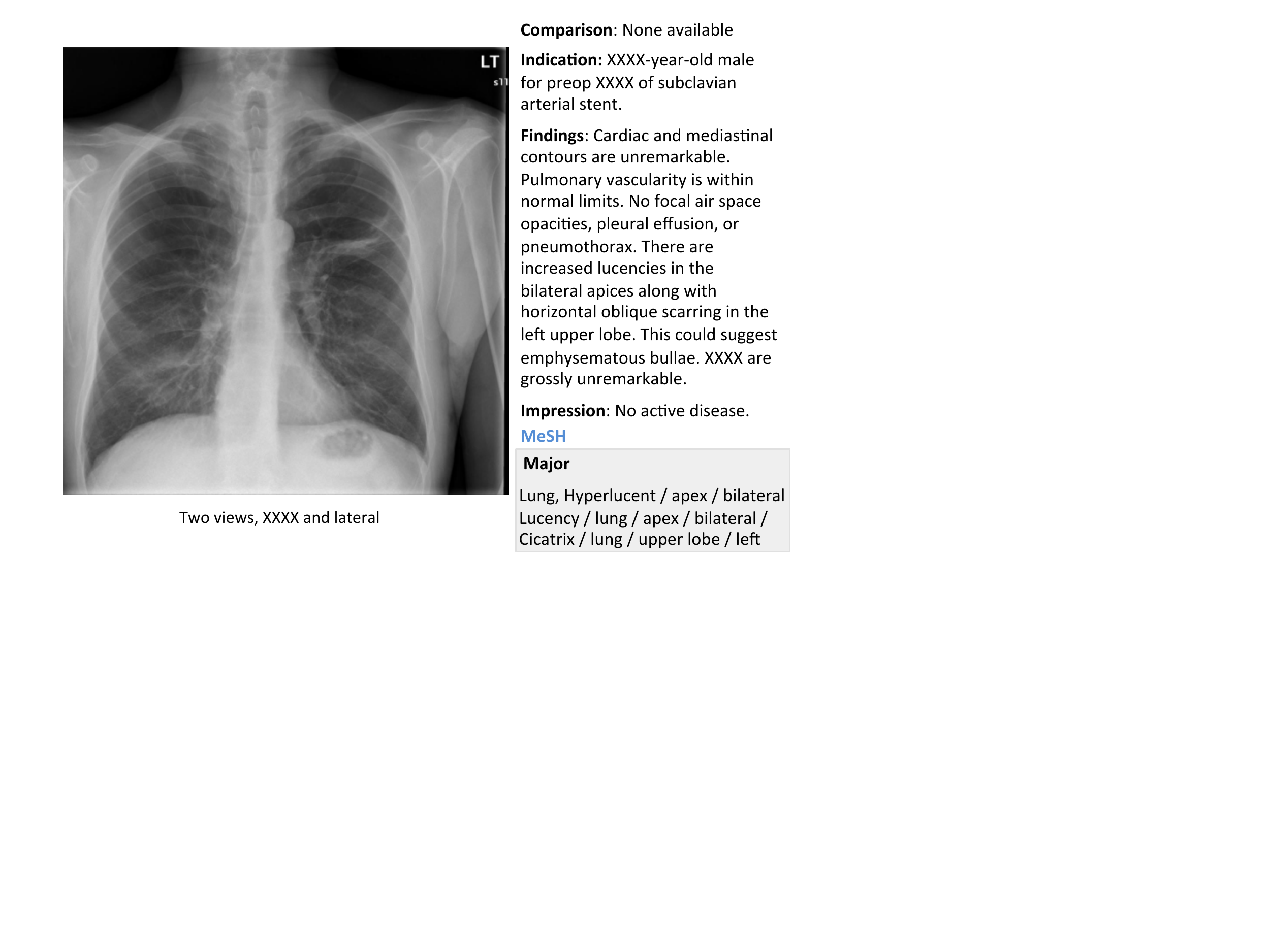}
\end{center}
   \caption{Examples of the OpenI \cite{openi} chest x-ray image, report, and annotations.}
\label{fig:chestx_example_02}
\end{figure}

More examples of chest x-ray image, report, and annotations available on OpenI \cite{openi} is shown in Figure \ref{fig:chestx_example_02}.\\

\section{Visualizations of Joint Image/Text Context Vectors}

\begin{figure*}[t]
\begin{center}
   \includegraphics[width=.78\linewidth]{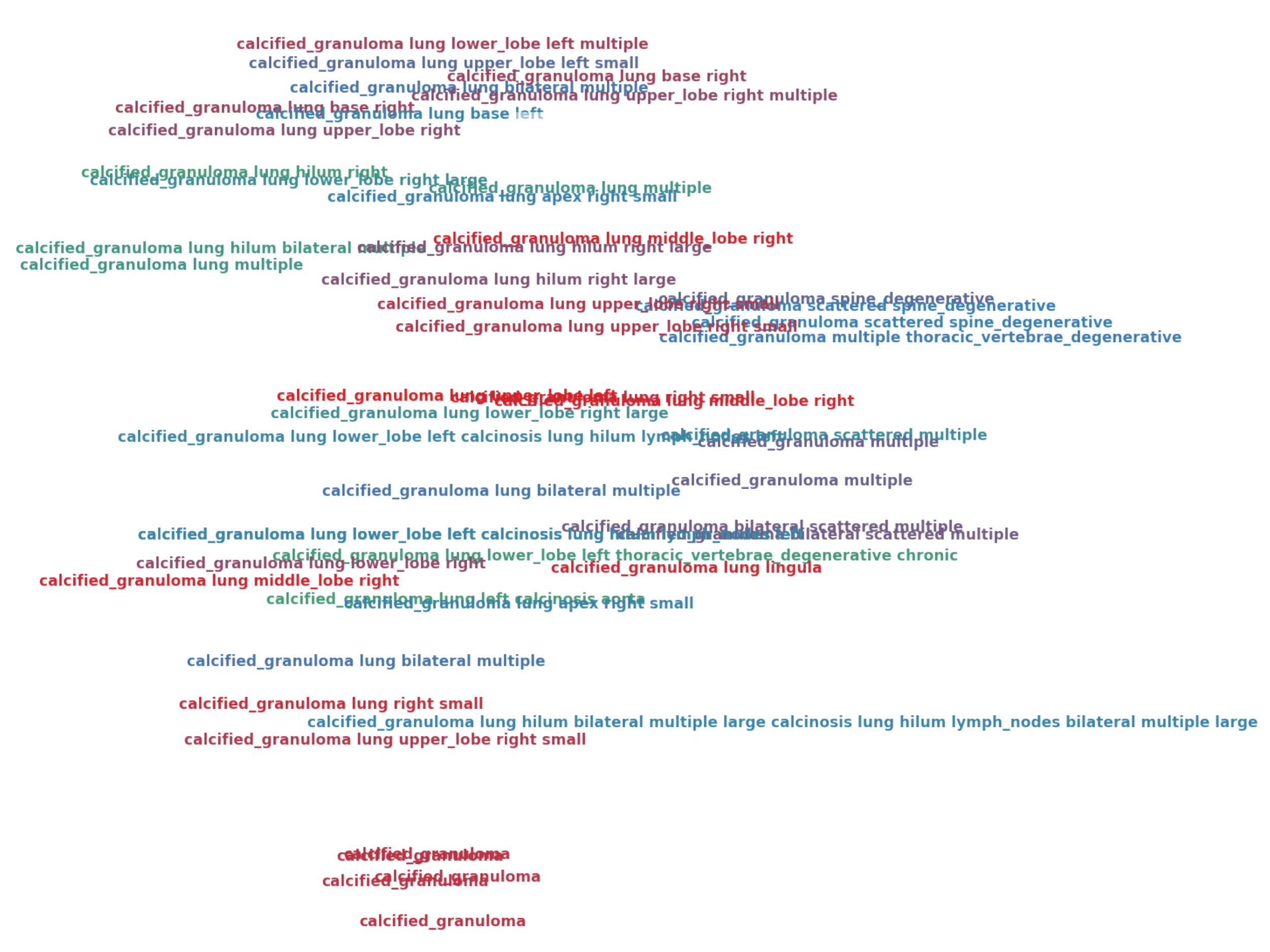}
\end{center}
   \caption{Visualization of the $h_\text{im:text}$ of about 50 samples of the annotations describing the disease \textit{calcified granuloma} on 2D plane.}
\label{fig:imtext_context_vector_viz_1_1}
\end{figure*}

\begin{figure*}[t]
\begin{center}
   \includegraphics[width=.75\linewidth]{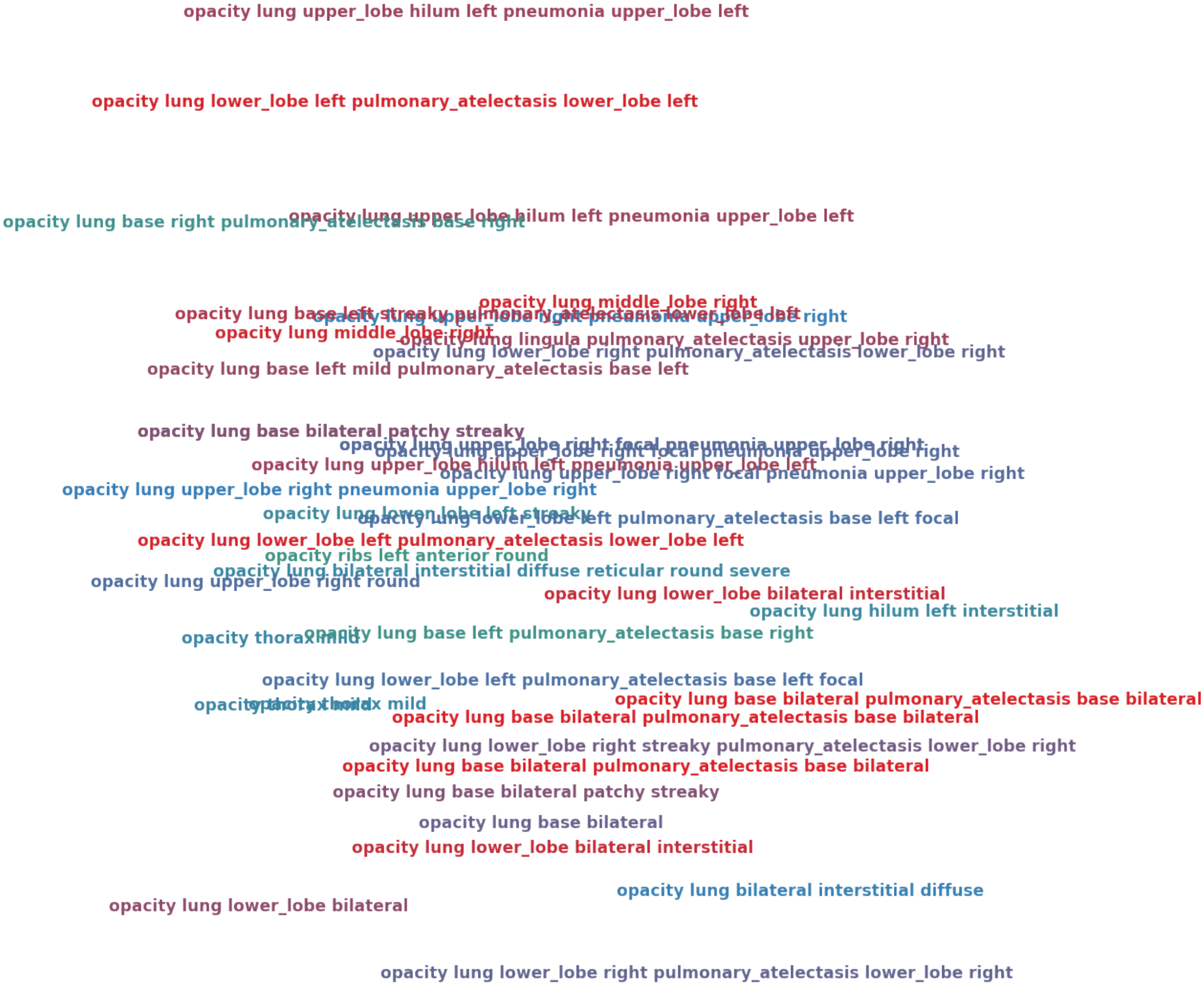}
\end{center}
   \caption{Visualization of the $h_\text{im:text}$ of about 50 samples of the annotations describing the disease \textit{opacity} on 2D plane.}
\label{fig:imtext_context_vector_viz_1_2}
\end{figure*}

Figure \ref{fig:imtext_context_vector_viz_1_1} shows the $h_\text{im:text}$ vectors of about fifty annotations describing \textit{calcified granuloma} are projected onto two-dimensional planes via dimensionality reduction ($\mathbb{R}^{1\times 1024} \rightarrow \mathbb{R}^{1\times 2}$), using the t-SNE \cite{van2008visualizing} implementation of \cite{scikit-learn}.
Figure \ref{fig:imtext_context_vector_viz_1_2} shows the visualization for the annotations describing \textit{opacity}.
We can see that different annotations describing a disease are thereby separated into different categories by the $h_\text{im:text}$.\\
\vspace{.25cm}

\section{More Annotation Generation Examples}

\begin{figure*}[t]
\vspace{-1.5cm}
\begin{center}
   \includegraphics[width=1.0\linewidth]{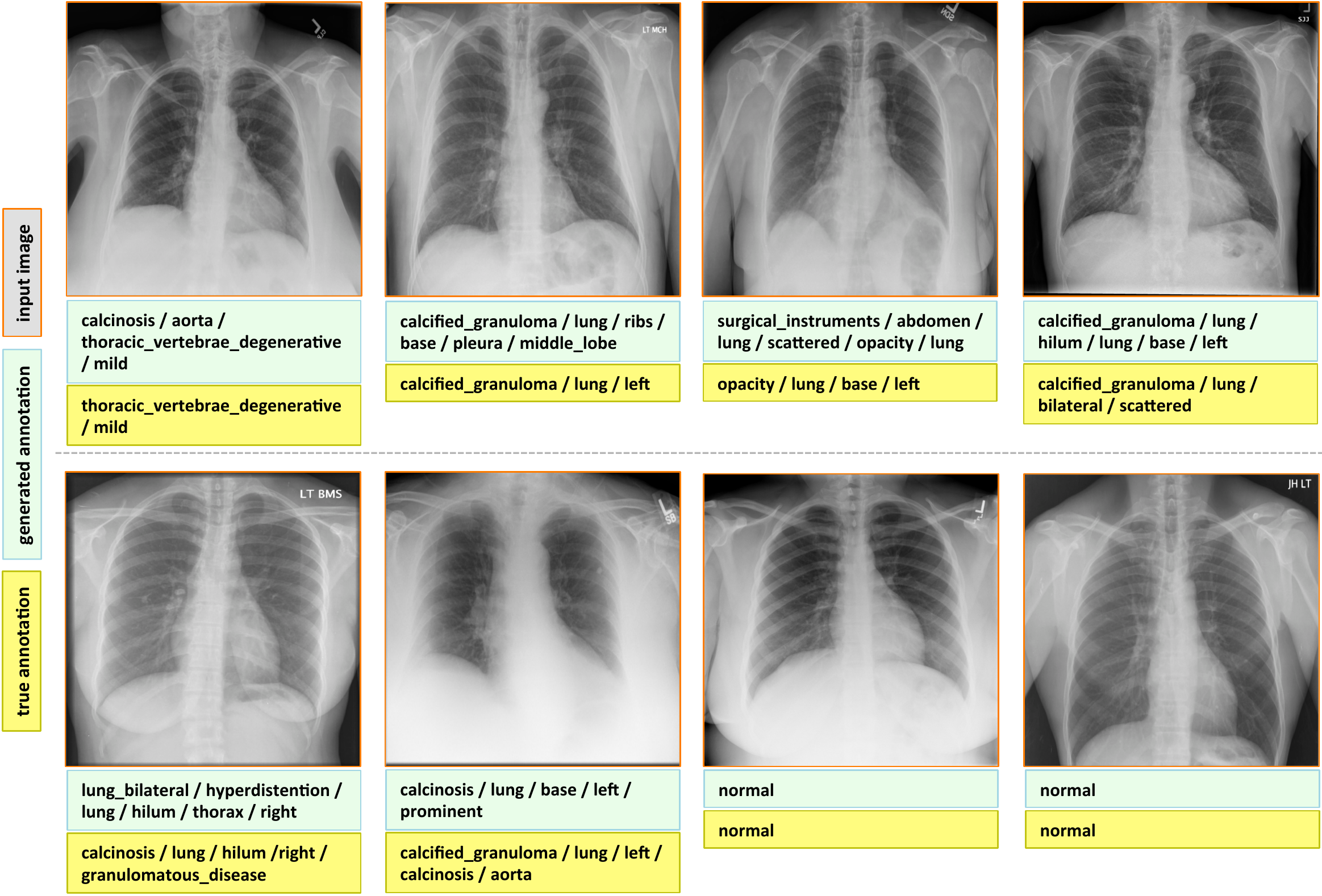}
   - - - - - - - - - - - - - - - - - - - - - - - - - - - - - - - - - - - - - - - - - - - - - - - - - - - - - - - - - - - - - - - - - - - - - - - - - - - - - - - - - - - - - - - - 
   \includegraphics[width=1.0\linewidth]{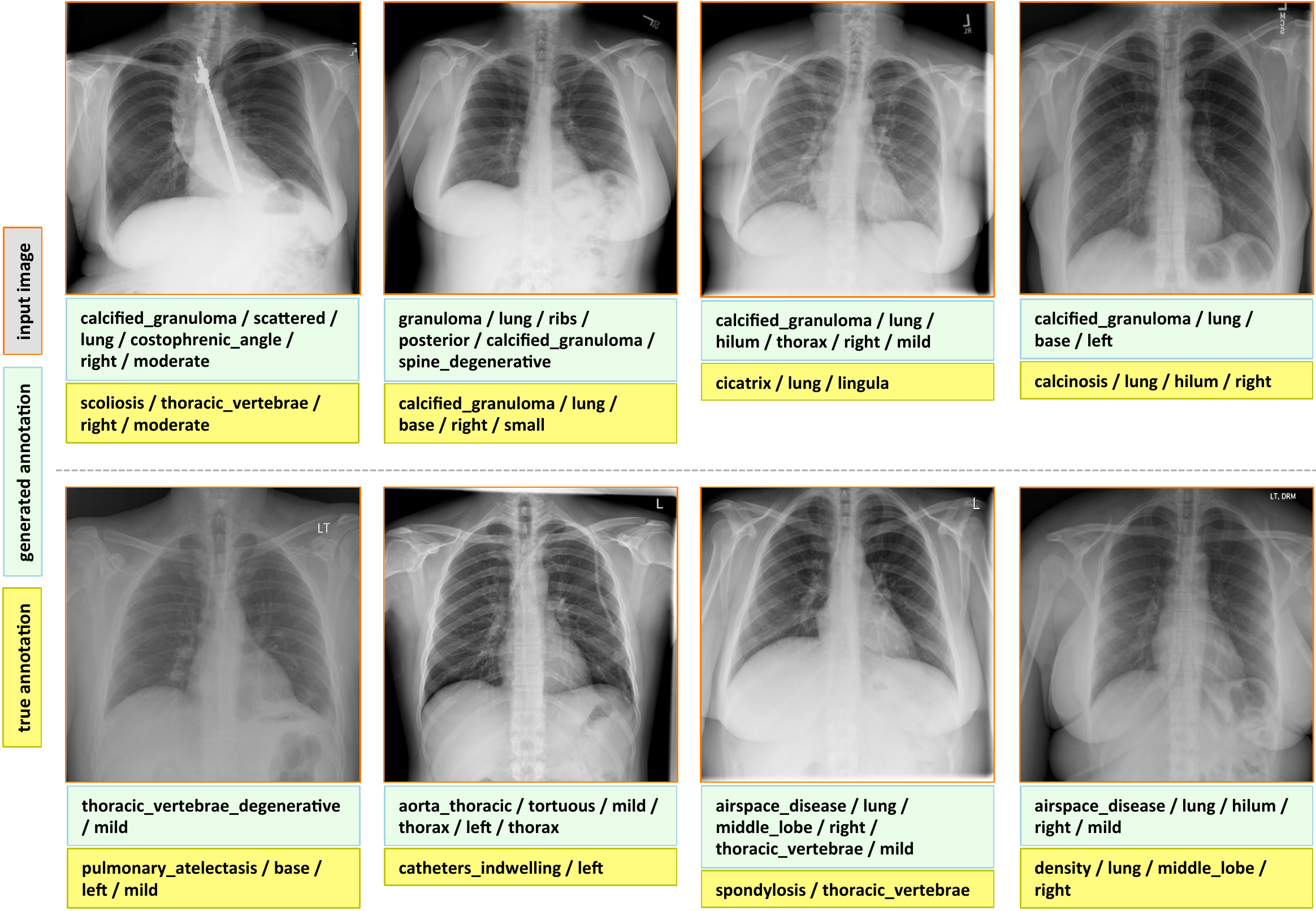}
\end{center}
   \caption{More examples of annotation generations (light green box) compared to true annotations (yellow box) for input images in the test set. Rare disease cases are more difficult to detect.}
\label{fig:annotation_generation_examples_3}
\end{figure*}

\begin{figure*}[t]
\vspace{-1.9cm}
\begin{center}
   \includegraphics[width=1.0\linewidth]{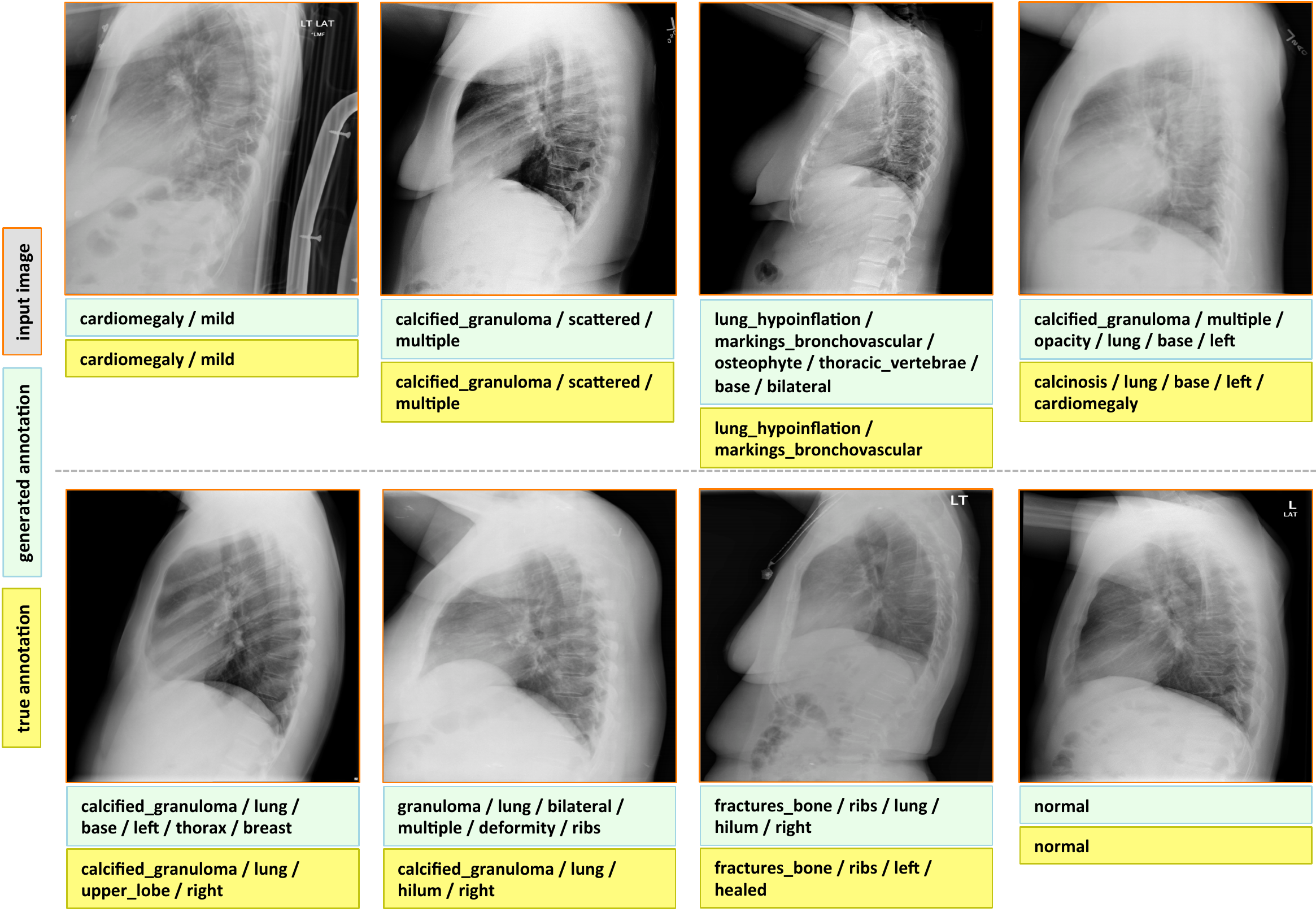}
   - - - - - - - - - - - - - - - - - - - - - - - - - - - - - - - - - - - - - - - - - - - - - - - - - - - - - - - - - - - - - - - - - - - - - - - - - - - - - - - - - - - - - - - - 
   \includegraphics[width=1.0\linewidth]{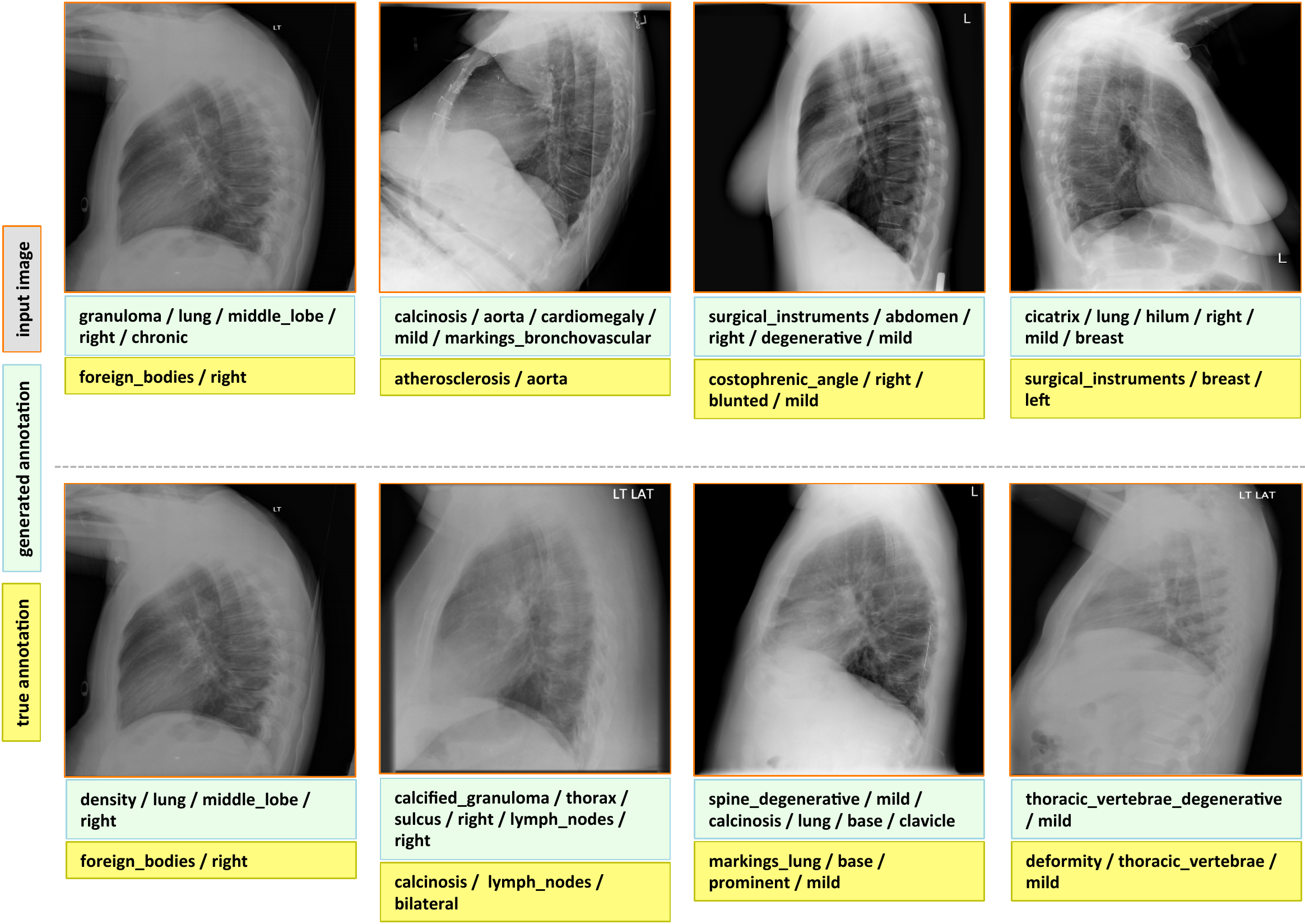}
\end{center}
   \caption{More examples of annotation generations (light green box) compared to true annotations (yellow box) for input images in the test set. In lateral (side) views, the (left or right) location of the disease can no longer be indentified.}
\label{fig:annotation_generation_examples_4}
\end{figure*}

More annotation generation examples are provided in Figures \ref{fig:annotation_generation_examples_3} and \ref{fig:annotation_generation_examples_4}.
Overall, the system generates promising results on predicting disease (labels) and its context (attributes) in the images.
However, rare disease cases are more difficult to detect.
For example, the cases \textit{pulmonary\_atelectasis}, \textit{spondylosis}, and \textit{density} (Figure \ref{fig:annotation_generation_examples_3}), as well as \textit{foreign\_bodies}, \textit{atherosclerosis}, \textit{costophrenic\_angle}, \textit{deformity} (Figure \ref{fig:annotation_generation_examples_4})  are much rarer in the data than \textit{calcified\_granuloma}, \textit{cardiomegaly}, and all the frequent cases listed in Table 1 of the main paper. 

Furthermore, the (left or right) location of the disease cannot be identified in a lateral view (obtained by scanning the patient from the side), as shown in Figure \ref{fig:annotation_generation_examples_4}.
Since our dataset contains a limited number of disease cases, we treat each x-ray image and report as a sample, and do not account for different views.

For future work, we plan to improve prediction accuracy by both (a) accounting for the different views, and (b) collecting a larger dataset to better account for rare diseases.

\vspace{0.5cm}

\end{appendices}

\end{document}